\documentclass[11pt,a4paper]{article}
\pdfoutput=1
\usepackage[hyperref]{acl2021}
\usepackage{times}
\usepackage{latexsym}
\usepackage{amsmath}

\usepackage{microtype}

\usepackage{graphicx}
\usepackage{blindtext}
\usepackage{multirow}
\usepackage{multicol}
\usepackage{float}

\usepackage{colortbl}
\definecolor{grey}{rgb}{0.8,0.8,0.8}

\aclfinalcopy 
\def\aclpaperid{1315} 

\title{How Reliable are Model Diagnostics?}

\author{Vamsi Aribandi\thanks{\hspace{1.5mm}Google AI Resident} \\
  Google Research \\
  \texttt{aribandi@google.com} \\
  \And
  Yi Tay \\
  Google Research \\
  \texttt{yitay@google.com} \\
  \And
  Donald Metzler \\
  Google Research \\
  \texttt{metzler@google.com} \\
  }

\date{}

\DeclareMathOperator{\mean}{mean}
\DeclareMathOperator{\stdev}{stdev}
\DeclareMathOperator{\pr}{Pr}

\begin{document}
\maketitle
\begin{abstract}
In the pursuit of a deeper understanding of a model's behaviour, there is recent impetus for developing suites of probes aimed at diagnosing models beyond simple metrics like accuracy or BLEU. This paper takes a step back and asks an important and timely question: how \emph{reliable} are these diagnostics in providing insight into models and training setups? We critically examine three recent diagnostic tests for pre-trained language models, and find that likelihood-based and representation-based model diagnostics are not yet as reliable as previously assumed. Based on our empirical findings, we also formulate recommendations for practitioners and researchers.
\end{abstract}

\section{Introduction}
Contemporary statistical models based on deep learning have made incredible progress towards solving complex language tasks \citep{radford2019language, devlin2019bert, JMLR:v21:20-074}. These models generally trade off the interpretability and simplicity of traditional models for powerful parameterizations and inductive biases, enabling their impressive performance. However, their entry into critical fields such as medicine, the justice system, and social media moderation often makes this trade-off a costly one. Consequently, there has been surging interest in the development of tools and suites for diagnosing and better understanding model behaviour, and gaining insight into what patterns and phenomena they have learned (\S\ref{sec:diagnostic_works}).

Ideally, these diagnostics would not only help practitioners understand the failure modes and capabilities of large contemporary models, but also enable them to improve their models based on the diagnostics. To this end, we believe that model diagnostics are essential for making meaningful progress in natural language processing.

Model diagnostics generally probe a model for specific learned qualities (\S\ref{sec:diagnostic_works}). These may be a positive qualities (e.g., whether a model has acquired syntactic knowledge) or potentially problematic qualities (e.g., biases and stereotypes. These probes can be used to identify certain phenomena that can be used to further improve models.

Given the potential impact that model diagnostics can have for practitioners and the research community's fundamental understanding of contemporary models, this paper asks the important and inevitable question of whether these probes are actually reliable and robust, and to what extent they are. These diagnostics' explicit nature as a tool for understanding also imposes a greater bar for robustness, as inconsistencies may mislead and result in compounding errors.

Our findings demonstrate that model diagnostics can be unreliable on multiple fronts. To illustrate our point, we select three diagnostics tasks --- StereoSet \citep{nadeem2020stereoset}, CrowS-Pairs \citep{nangia-etal-2020-crows}, and SEATs \citep{may-etal-2019-measuring} to base our empirical evaluation on. 
Overall, we find that \textbf{likelihood-based and representation-based diagnostics measured multiple times on the same training setup can result in wildly different findings.} Specifically, a substantial variance is observed when performing the same model diagnostics on identical BERT \cite{devlin2019bert} pre-training setups while varying minute details such as the initial random seed or choice of representation.

These findings are meant to caution researchers and practitioners that rely on such diagnostics so that they can be more mindful of these phenomena when analyzing their models in the future. We discuss the implications of our findings and propose recommendations for practitioners and researchers in \S\ref{sec:discussion}.

\section{Methodology}
\label{sec:methodology}

\subsection{Training setup}
\label{sec:setup}
We pre-train 5 BERT \textsc{base} and \textsc{large} uncased English models, each with the same configurations as in \citet{devlin2019bert} using Tensorflow\footnote{\url{https://github.com/tensorflow/models/tree/master/official/nlp/bert}}. However, each model differs in its random seed, resulting in different parameter initializations and training data permutations. Hence, it is expected that the checkpoints will each end up at a different local minima. It should be noted that BERT uses static masking instead of dynamic masking, so the set of pre-training examples remains the same.

To decouple our findings from phenomena that occur as a result of using different training setups, we restrict our experiments to only those that require pre-trained BERT models, eliminating many probes mentioned in \S\ref{sec:underspec_works}. \citet{webster2020measuring} report that patterns learned during pre-training are often resilient to fine-tuning, further supporting our reasoning.

\subsection{Likelihood-base diagnostics}
\label{sec:likelihoods}

One approach to examining the behaviour of language models like BERT is to examine how they rank certain representative examples above others. We use two contemporary datasets that measure how often stereotypes are ranked above anti-stereotypes --- StereoSet \citep{nadeem2020stereoset} and CrowS-Pairs \citep{nangia-etal-2020-crows}. Both datasets measure
$
\mathit{ss} = 100*\sum_{n=1}^{|X|}1_{[\mathit{ll}(x_{n}^\mathit{ster}) > \mathit{ll}(x_{n}^\mathit{anti})]}/|X|
$.

\paragraph{StereoSet}
\citet{nadeem2020stereoset} propose a benchmark that contains intra-sentence and inter-sentence examples of stereotypes and anti-stereotypes. Here, likelihoods are calculated as $\mathit{ll}(x)=p(x_\tau|x_{\setminus\tau})$ (where $\tau$ is the set of target demographic word(s) in $\mathit{x}$) and $\mathit{ll}(x)=p(\verb|isNext||x_1, x_2)$ for intra-sentence and inter-sentence examples respectively.
They also propose and combine a language modeling score ($\mathit{lms}$) with $\mathit{ss}$ into a hybrid metric ($\mathit{icat}$), but we only report $\mathit{ss}$ to focus on StereoSet's primary purpose --- measuring stereotypical preference in language models. We report results on the development set.

\paragraph{CrowS-Pairs}
\label{sec:crows}
\citealp{nangia-etal-2020-crows} propose a test that contains intra-sentence examples, where likelihoods are calculated by conditioning on the target demographic word(s) in the sentence ($\mathit{ll}(x) = p(x_{\setminus\tau}|x_\tau)$) rather than vice-versa as in StereoSet.

The CrowS-Pairs diagnostic is expected to show higher variance than StereoSet for two reasons: (1) it is a smaller dataset (${\sim}\frac{1}{3}$rd the size of StereoSet-dev) with more categories, so results are more sensitive to changes in individual predictions; and (2) the pseudo-likelihood it uses is more susceptible to the poor calibration \citep{jiang2020know, desai-durrett-2020-calibration} of contemporary models, since the number of multiplied probabilities grows linearly with the number of words in a sentence.

\subsection{Vector-space diagnostics}
\label{sec:orientations}

Directly examining representations learned by models is another way to understand their behavior. This is typically done by measuring relationships between different types of inputs, for example in terms of their relative orientations in a vector space.

\paragraph{SEATs}
We use Sentence Encoder Association Tests (SEATs;~\citealp{may-etal-2019-measuring}), which extend the popular Word Embedding Association Tests (WEATs;~\citealp{Caliskan183}) by constructing ``semantically bleached'' sentences. A WEAT/SEAT measures the \emph{effect size} $s(X,Y,A,B)$ of the association between two targets (e.g., $X$=\verb|MentalDisease| and $Y$=\verb|PhysicalDisease|) and two attributes (e.g., $A$=\verb|Temporary| and $B$=\verb|Permanent|), as well as the statistical significance of the association using a permutation test\footnote{Please see Appendix \ref{appendix:seat} for how SEATs are computed.}. We conduct experiments using the same SEATs as in \citet{may-etal-2019-measuring}. In addition to testing sentence (\verb|[CLS]|) representations, we also test the contextualized word representations of the target/attribute words in the sentences. The reason we do this is that even for semantically bleached sentences, it is often non-trivial for models to encode information about an entire sentence in a single vector\footnote{\url{https://www.cs.utexas.edu/~mooney/cramming.html}}.

In addition to examining effect sizes, we also conduct an experiment to see how distinguishable representations of certain concepts are in vector space (e.g., do representations of \verb|Pleasant| and \verb|Unpleasant| sentences form their own clusters?). We do this by clustering (via \emph{k}-means) sentence representations and subsequently examining how well the unsupervised clusters align with the actual categories. The aim of this experiment is to understand vector space diagnostics behave the way they do.

\begin{figure}
    \centering
    \includegraphics[scale=0.5]{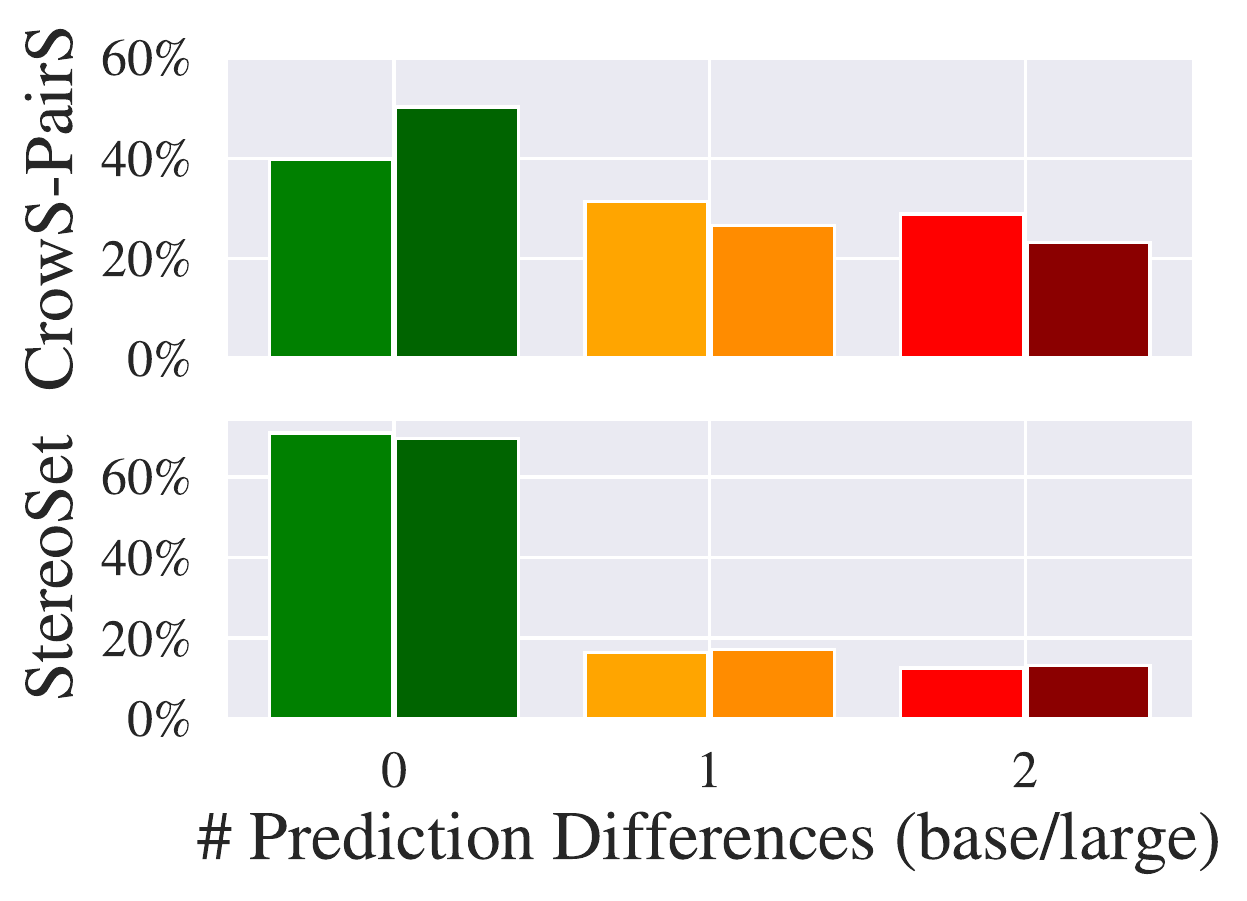}
    \caption{\% of examples in likelihood-based tests that have $d$ different predictions over 5 runs. Ideally, examples would always (100\%) be predicted the same ($d$=0).}
    \label{fig:ster_crow_diff}
\end{figure}

\begin{table}
\centering
\begin{tabular}{cllcc}
\hline \textbf{Test} & \textbf{Cat.} & \textbf{N} & \multicolumn{2}{c}{\textbf{BERT results (\%)}} \\
\cline{4-5}
& & & \textbf{\textsc{base}} & \textbf{\textsc{large}} \\ \hline
\multirow{9}{*}{\rotatebox[origin=c]{90}{CrowS-Pairs}}
& Race & 516 & \cellcolor{grey}$54.4 \pm \textbf{4.7}$ & $55.9 \pm \textbf{2.7}$ \\
& Gen. & 262 & $58.2 \pm 2.5$ & $61.1 \pm 1.7$ \\
& S.O. & 84 & $63.2 \pm 3.4$ & $67.4 \pm \textbf{4.6}$ \\
& Rel. & 105 & $68.9 \pm \textbf{8.0}$ & $72.2 \pm 2.1$ \\
& Age & 87 & \cellcolor{grey}$55.4 \pm \textbf{4.2}$ & $60.9 \pm \textbf{5.6}$ \\
& Nat. & 159 & $51.2 \pm 1.2$ & \cellcolor{grey}$55.3 \pm \textbf{3.5}$ \\
& Dis. & 60 & $69.0 \pm \textbf{3.8}$ & $79.0 \pm 1.9$ \\
& P.A. & 63 & $59.1 \pm \textbf{4.9}$ & $64.4 \pm \textbf{4.3}$ \\
& Occ. & 172 & \cellcolor{grey}$54.9 \pm \textbf{4.5}$ & $58.0 \pm \textbf{4.2}$ \\
\cline{2-5}
& all & 1508 & $57.1 \pm \textbf{2.8}$ & $60.3 \pm 1.7$ \\
\hline
\multirow{5}{*}{\rotatebox[origin=c]{90}{StereoSet}}
& Gen. & 496 & $59.1 \pm 0.7$ & $62.4 \pm 2.0$ \\
& Occ. & 1636 & $60.5 \pm 0.6$ & $61.4 \pm 0.8$ \\
& Race & 1938 & $54.8 \pm 1.1$ & $56.4 \pm 0.8$ \\
& Rel. & 156 & $51.8 \pm \textbf{2.8}$ & \cellcolor{grey}$54.4 \pm \textbf{3.3}$ \\
\cline{2-5}
& all & 4226 & $57.4 \pm 0.7$ & $59.0 \pm 0.7$ \\
\hline

\end{tabular}
\caption{\label{table:ster_crows} Likelihood-based diagnostics over categories often have high standard deviation (bold) over pre-training runs, often varying from almost neutral (${\sim}50\%$) to a significant amount (highlighted).}
\end{table}

\begin{figure}
    \centering
    \includegraphics[width=\linewidth]{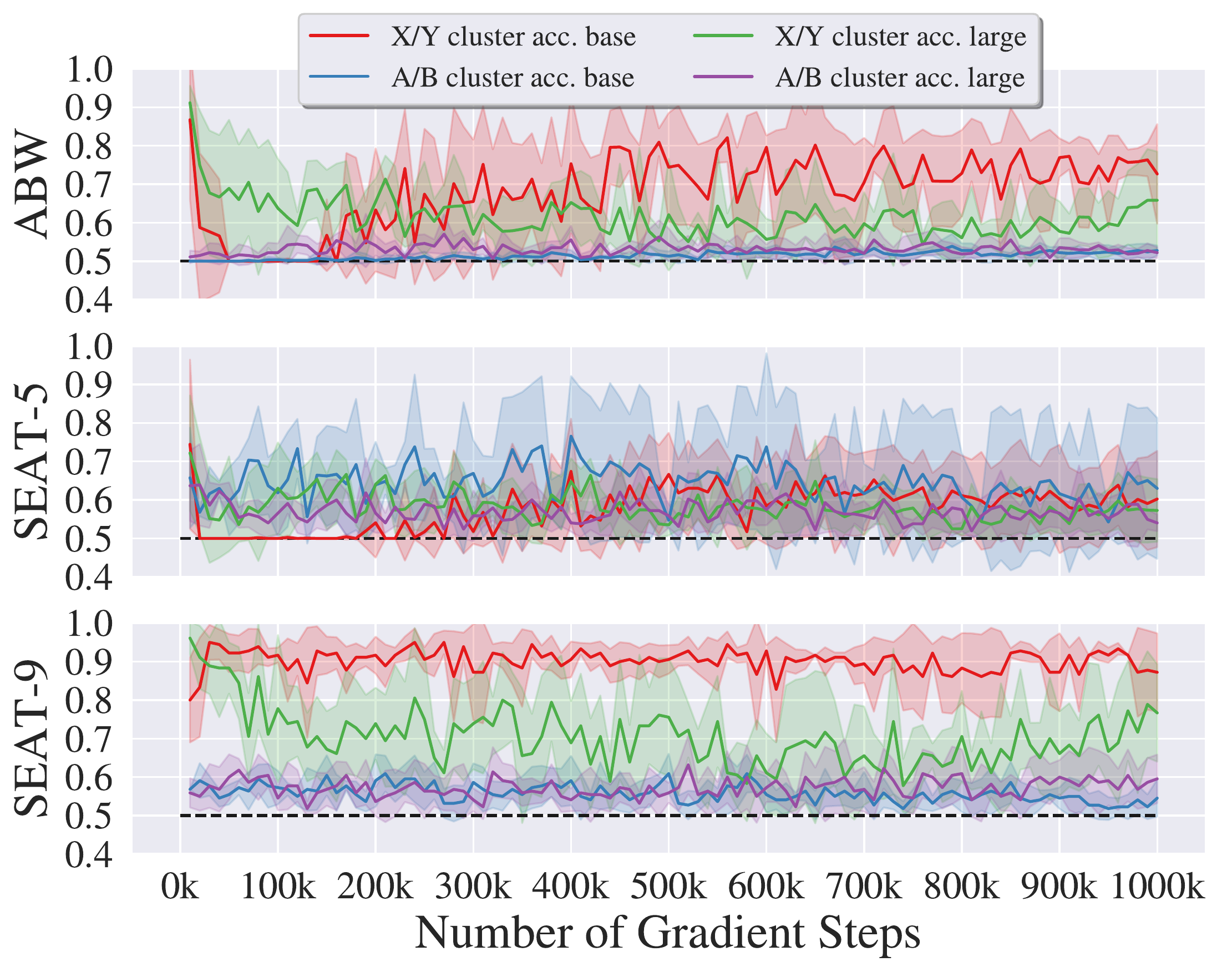}
    \caption{Representations of targets and/or attributes often don't cluster over pre-training. The dashed line is when representations are indistinguishable (acc. = 0.5).}
    \label{fig:seat_clu}
\end{figure}

\section{Findings and Insights}
\label{sec:findings}
\subsection{Likelihood-based diagnostics are unstable}
\label{sec:likelihood_findings}

Experiments on StereoSet and CrowS-Pairs show that while likelihood-based ranking diagnostics may be stable across all categories, instability is evident in the results of individual categories (Table \ref{table:ster_crows}). Many categories have a standard deviation of over 2.5 percentage points. Some categories also vary from almost no stereotypical preference to a significant amount (highlighted in Table \ref{table:ster_crows}) --- a result that could potentially cause practitioners to draw false conclusions.

Additionally, from Figure~\ref{fig:ster_crow_diff} it is evident that many examples are assigned different labels over the 5 pre-trained models, often having 3 models assign them one label and 2 models assigning them the opposite label --- almost as random as a coin flip! The implies that the models are probably uncertain about their predictions for these datapoints, motivating the consideration of model uncertainty in diagnostic measures instead of simply making a binary decision by comparing likelihoods.

Worryingly, both tests report wildly differing results on religious stereotypes (``Rel.''), with CrowS-Pairs detecting strong stereotypical preference and StereoSet detecting almost none. It is also worth noting that results on CrowS-Pairs exhibit far higher variance compared to StereoSet (Table \ref{table:ster_crows}, Figure \ref{fig:ster_crow_diff}), as hypothesized in \S\ref{sec:crows}.

\begin{figure*}
    \centering
    \includegraphics[width=\linewidth]{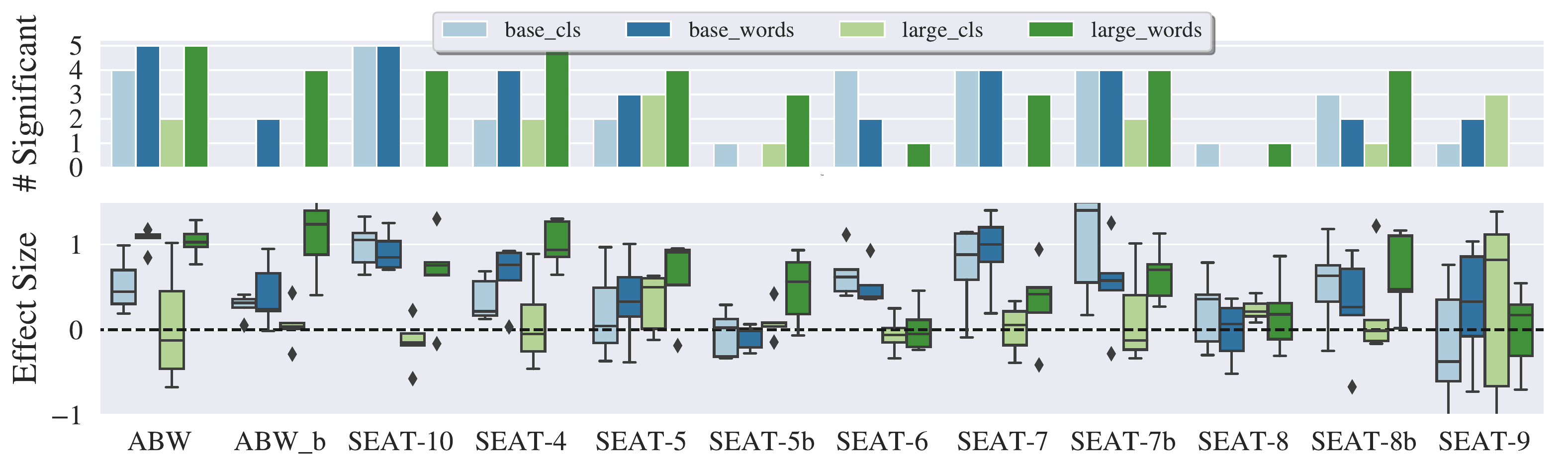}
    \caption{SEAT results exhibit high variance across pre-train runs, model sizes, and choice of representation. Moreover, effect sizes often vary around the ``neutral'' mark (0) and also have different statistical significances (at $p=0.01$). Ideally, a test would always (5) or never (0) be significant, and yield effect sizes with the same sign.}
    \label{fig:seat_bars}
\end{figure*}

\subsection{Vector-space diagnostics are unstable}
\label{sec:vecspace_findings}
Representation-based experiments exhibit high variance across multiple pre-training runs, choices of representation, and model sizes (Figure~\ref{fig:seat_bars}). Notably, SEAT results are often on both sides of the ``neutral'' mark (0), and their statistical significance is often erratic. In other words, it is possible for two models to be pre-trained with the exact same configurations but different random seeds to yield completely opposite conclusions on some SEATs. Moreover, the same checkpoint often yields different results depending on whether sentence or pooled target-word representations are used. Ideally, a SEAT would \emph{always} or \emph{never} be statistically significant, and yield effect sizes with the same sign over multiple pre-training runs and (seemingly innocuous) choices of representation.

From Figure~\ref{fig:seat_clu}, the representational instability of semantically bleached SEAT sentences is further evident --- how these representations cluster together is erratic both across pre-training steps as well as across multiple pre-training runs. This result gives us further insight into why high variance is observed for vector-space diagnostics --- representations often can't form their own clusters for certain concepts, so simply examining their relative orientations is insufficient. Our findings provide empirical arguments for what \citet{may-etal-2019-measuring} surmise --- there is scope for sentence embedding-based tests that do more than naturally extend word embedding-based tests with semantically bleached sentences.

We surmise that representation-based diagnostics are less stable than likelihood-based diagnostics because large models like BERT are optimized to be good at modeling likelihoods via their pre-training objective. However, there is no constraint on how sentences must be represented other than it should be possible to ``extract" correct likelihoods from them. In other words, there is no reason to expect the orientations of these representations to provide deep insight into what these models learn.

\subsection{Diagnostic instability is despite equivalent downstream performance}
We fine-tuned the 10 checkpoints on SST-2 \citep{socher2013recursive}, RTE \citep{dagan2006pascal, bar2006second, giampiccolo2007third, bentivogli2009fifth}, and QNLI \citep{rajpurkar2016squad} from the GLUE benchmark \citep{wang2019glue}. Development-split results show that performance was largely the same across checkpoints (Table \ref{tab:finetuning}) despite diverging behaviour on the model diagnostics as shown in \S\ref{sec:likelihood_findings} and \S\ref{sec:vecspace_findings}. \textbf{This shows that the different local optima still perform largely the same on downstream tasks despite behaving differently with respect to model diagnostics}.

Dev-set performance is also largely consistent with what is expected of BERT \textsc{base} and \textsc{large} models. It should be noted that we only used one set of hyperparameters and did not perform the hyperparameter sweep as in \citet{devlin2019bert}, so further tuning would likely improve results.

\begin{table}
\centering
\begin{tabular}{lcc}
\hline \textbf{Dataset} & \multicolumn{2}{c}{\textbf{BERT fine-tuning results}} \\
\cline{2-3}
& \textbf{\textsc{base}} & \textbf{\textsc{large}} \\ \hline
SST-2 & $91.2 \pm 0.3$ & $93.0 \pm 0.3$ \\
RTE & $71.3 \pm 1.2$ & $76.8 \pm 1.8$ \\
QNLI & $92.1 \pm 0.2$ & $92.1 \pm 0.3$ \\

\end{tabular}
\caption{\label{tab:finetuning} The checkpoints generally exhibit equivalent performance on downstream tasks.}
\end{table}

\section{Related Work}
\label{sec:related}
\subsection{Model Diagnostics}
\label{sec:diagnostic_works}
Models have been probed to understand what exactly they learn beyond traditional language tasks, ranging from their linguistic capabilities \citep{adi2017finegrained, tenney2018what, conneau-etal-2018-cram, ribeiro-etal-2020-beyond, belinkov-etal-2017-neural, hewitt-manning-2019-structural, marvin-linzen-2018-targeted}, multilingual capabilities \citep{pires-etal-2019-multilingual, kudugunta-etal-2019-investigating}, world knowledge \citep{jiang-etal-2020-know, petroni-etal-2019-language}, and social bias \citep{nadeem2020stereoset, nangia-etal-2020-crows, may-etal-2019-measuring} among other phenomena.

Another axis to compare model diagnostics on is whether they are intrinsic or extrinsic, i.e., whether they directly analyze models for certain phenomena that aren't tied to any downstream task or do so keeping particular tasks in mind. This paper restricts itself to intrinsic tasks for reasons mentioned in \S\ref{sec:setup}. An example of an extrinsic task is \citet{rudinger-etal-2018-gender}, which probes models for gender bias through the lens of coreference resolution. We refer readers to \citet{belinkov-glass-2019-analysis} for a more comprehensive survey on model analysis for natural language processing.

\subsection{Diagnostic Fragility}
\label{sec:fragility_works}
It has been shown that classifier probes --- which require an additional classifier (like an MLP) to be trained on top of frozen model representations --- are unstable \citep{voita-titov-2020-information}, and that it might not be clear from their results whether the probe \emph{itself} learned a phenomena or whether the diagnosed representations learned it \citep{hewitt-liang-2019-designing}. Similarly, \citet{wang-etal-2020-gradient} find that gradient-based analysis of language technologies based on neural networks can often be unreliable and manipulable. Attention-based interpretation can also be unreliable and manipulable to the point of deceiving practitioners, as \citet{pruthi-etal-2020-learning} and \citet{jain-wallace-2019-attention} show. The works mentioned above all support our arguments, and some raise similar concerns to those expressed in this paper.

\subsection{Inconsistencies between equivalent checkpoints}
\label{sec:underspec_works}
This paper's findings can be linked to the problems caused by underspecification in machine learning \citep{damour2020underspecification}, i.e., when multiple unique predictors trained with the same configuration have the same performance but differ in subtle ways. In a setting where practitioners might train and thoroughly analyze one model but then retrain it and assume that the first checkpoint's model diagnostics hold for the second one, this issue is highly relevant. \citet{mccoy-etal-2020-berts} also find that separately fine-tuned BERT models often vary significantly in generalizing to auxiliary tasks.

\section{Discussion}
\label{sec:discussion}
\paragraph{Recommendations}
No probe is perfect, but it is clear that model diagnostics are not as reliable as previously assumed. Our empirical findings --- coupled with the works mentioned in \S\ref{sec:fragility_works} and \S\ref{sec:underspec_works} --- motivate careful scrutiny of model diagnostics. \textbf{We recommend that:}

\begin{itemize}
    \vspace{-2mm}
    \item Practitioners not generalize a single diagnostic result to the entire training setup, and instead restrict conclusions to a specific checkpoint.
    \vspace{-2mm}
    \item Researchers proposing probes not only test on publicly available checkpoints, but rather examine a probe's performance and robustness across a range of model/probe configurations.
\end{itemize}

\paragraph{Future Work}
While this paper primarily aims to motivate further scrutiny of model diagnostics, we hope it motivates studies that ask \emph{why} these diagnostics often behave unreliably. One future research direction we are excited about is analyzing correlations between the properties of the models' local minima in the loss landscape and behaviour on model diagnostics. This would not only be another step towards a better understanding of how contemporary deep language models work, but also enable researchers to use that information to design better, more robust model diagnostics. Such a study may even help inform the optimization process for future state-of-the-art language technologies.

It should also be noted that this paper is restricted to three diagnostics spanning likelihood-based and representation-based probes, and that future work is needed to determine the extent to which other diagnostic probes are reliable.

\section{Conclusion}
In this paper, we motivate further scrutiny of model diagnostics that aim to understand the behaviour of contemporary ``black-box'' language technologies. Our results show that model diagnostics are often fragile and can yield different conclusions as a result of seemingly innocuous configuration changes. We hope that our results over multiple pre-train runs will encourage researchers and practitioners to be mindful of the reliability of such model diagnostics when verifying hypotheses about their models and training setups.

\bibliographystyle{acl_natbib}
\bibliography{anthology,acl2021}

\appendix

\pagebreak

\section{SEAT computation}
\label{appendix:seat}
The effect size of a SEAT --- characterized by two target ($X, Y$) and two attribute ($A, B$) sets of sentences --- is calculated as:
$$
d = \frac{\mean_{x \in X}{s(x, A, B)} - \mean_{y \in Y}{s(y, A, B)}}{\stdev_{z \in X \cup Y}s(z, A, B)}
$$
where:
\begin{align*}
s(\mathit{sent}, A, B) = \mean_{a \in A}\cos(\overrightarrow{sent}, \overrightarrow{a})\\
- \mean_{b \in B}\cos(\overrightarrow{sent}, \overrightarrow{b}).
\end{align*}
The \textit{p}-value of the permutation test to determine the statistical significance of the effect size is calculated as::
$$
p = \pr[S(X_i, Y_i, A, B) > S(X, Y, A, B)]
$$
over partitions $(X_i, Y_i)$ of $(X \cup Y)$ such that $|X_i|=|Y_i|$, where:
$$
S(X,Y,A,B) = \sum_{x \in X}{s(x, A, B)} - \sum_{y \in Y}{s(y, A, B)}
$$

\end{document}